\documentclass{INTERSPEECH2023}
\usepackage{amsmath,graphicx}
\usepackage{bm}
\usepackage{multirow}
\usepackage{float}
\usepackage{makecell}

\usepackage{caption} 
\usepackage{hyperref} 
\usepackage{algorithm}
\usepackage{algorithmic}

\interspeechcameraready

\title{Rethinking Speech Recognition with A Multimodal Perspective via Acoustic and Semantic Cooperative Decoding}

\name{
    \begin{tabular}{c}
       \large ~Tian-Hao Zhang$^{1, 2}$, ~Hai-Bo Qin$^{1}$, ~Zhi-Hao Lai$^{1, 2}$,
       ~Song-Lu Chen$^{1, 2}$, ~Qi Liu$^{1, 2}$, \\
       ~Feng Chen$^{2, 3}$, ~Xinyuan Qian$^{1, *}$, ~Xu-Cheng Yin$^{1, 2}$\thanks{* Corresponding author.}
   \end{tabular}
    \vspace{-10pt}
}
\address{
    $^1$\normalsize School of Computer and Communication Engineering, University of Science and Technology Beijing, Beijing 100083, China \\
    $^2$ \normalsize USTB-EEasyTech Joint Lab of Artificial Intelligence, University of Science and Technology Beijing, Beijing 100083, China\\
    $^3$ \normalsize EEasy Technology Company Ltd., Zhuhai 519000, China
}
\email{\{tianhaozhang, zhihaolai\}@xs.ustb.edu.cn, \{songluchen, xinyuanqian\}@ustb.edu.cn}

\begin{document}

\maketitle
 
\begin{abstract}
Attention-based encoder-decoder (AED) models have shown impressive performance in ASR. However, most existing AED methods neglect to simultaneously leverage both acoustic and semantic features in decoder, which is crucial for generating more accurate and informative semantic states. In this paper, we propose an Acoustic and Semantic Cooperative Decoder (ASCD) for ASR. In particular, unlike vanilla decoders that process acoustic and semantic features in two separate stages, ASCD integrates them cooperatively. To prevent information leakage during training, we design a Causal Multimodal Mask. Moreover, a variant Semi-ASCD is proposed to balance accuracy and computational cost. Our proposal is evaluated on the publicly available AISHELL-1 and aidatatang\_200zh datasets using Transformer, Conformer, and Branchformer as encoders, respectively. The experimental results show that ASCD significantly improves the performance by leveraging both the acoustic and semantic information cooperatively.

\end{abstract}

\section{Introduction}
Automatic speech recognition (ASR) has gained significant advances in recent years, largely driven by the development of deep learning models \cite{ctc, graves2012sequence, asr1, asr2, aed1}. Among these models, the attention-based encoder-decoder (AED) architecture has emerged as a powerful methods for improving ASR performance. Specifically, AED combines an encoder network that processes acoustic features with a decoder network that produces text transcriptions. The attention mechanism \cite{vaswani2017attention} allows the decoder to selectively focus on different parts of the acoustic features, enabling the model to better capture long-range dependencies, which eventually improves the ASR accuracy.

\begin{figure}[t]
  \centering
    \setlength{\abovecaptionskip}{-1.mm}
  \includegraphics[scale=0.31]{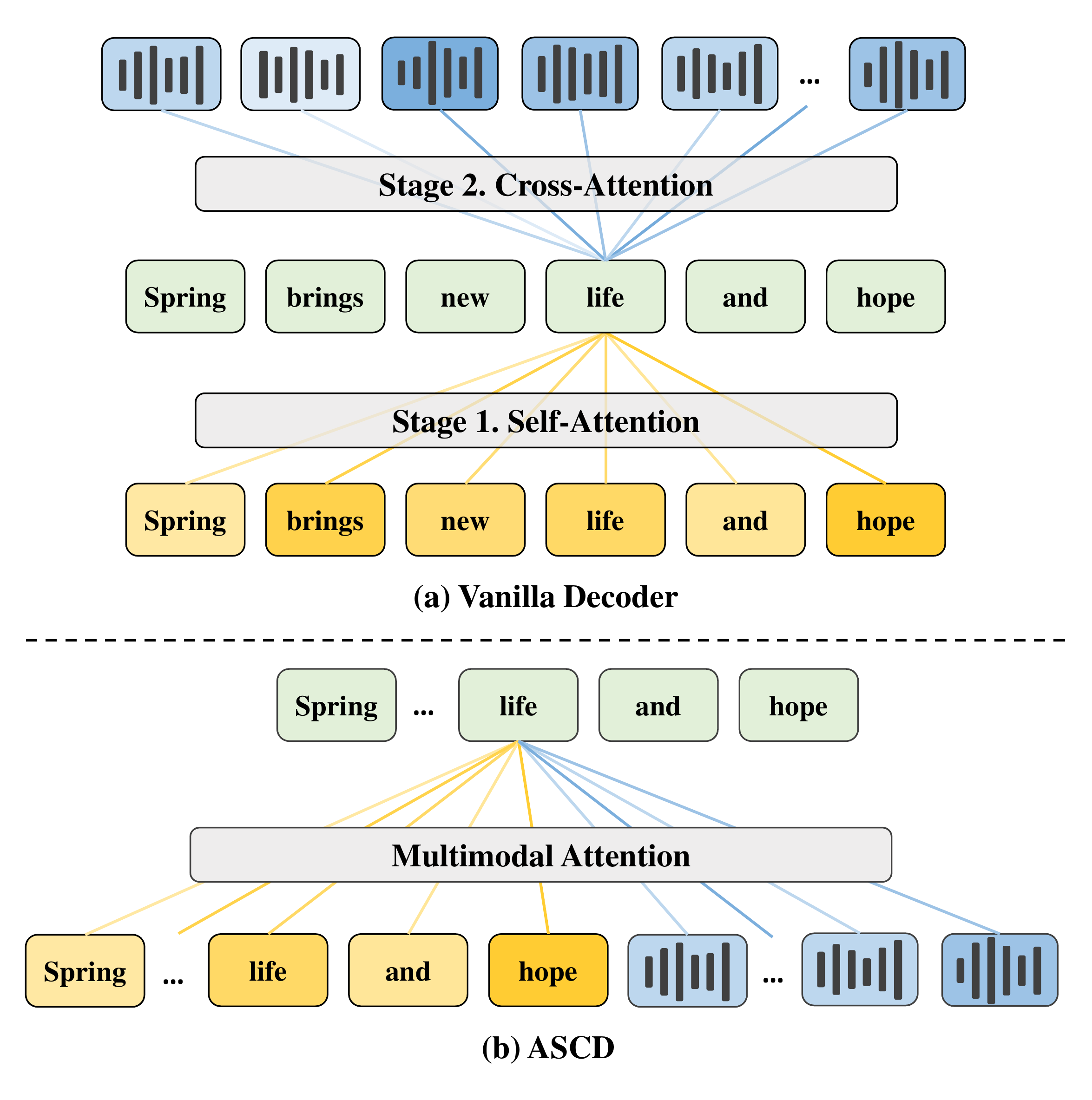}

  \caption{The distinct interaction processes of audio and text features in (a) the Vanilla Decoder and (b) our proposed ASCD. The shading of the color blocks and lines represents the attention weight assigned to each feature. }
  \label{fig:model}
    \vspace{-0.7cm}
\end{figure}

\begin{figure*}[ht]
  \centering
  \includegraphics[scale=0.7]{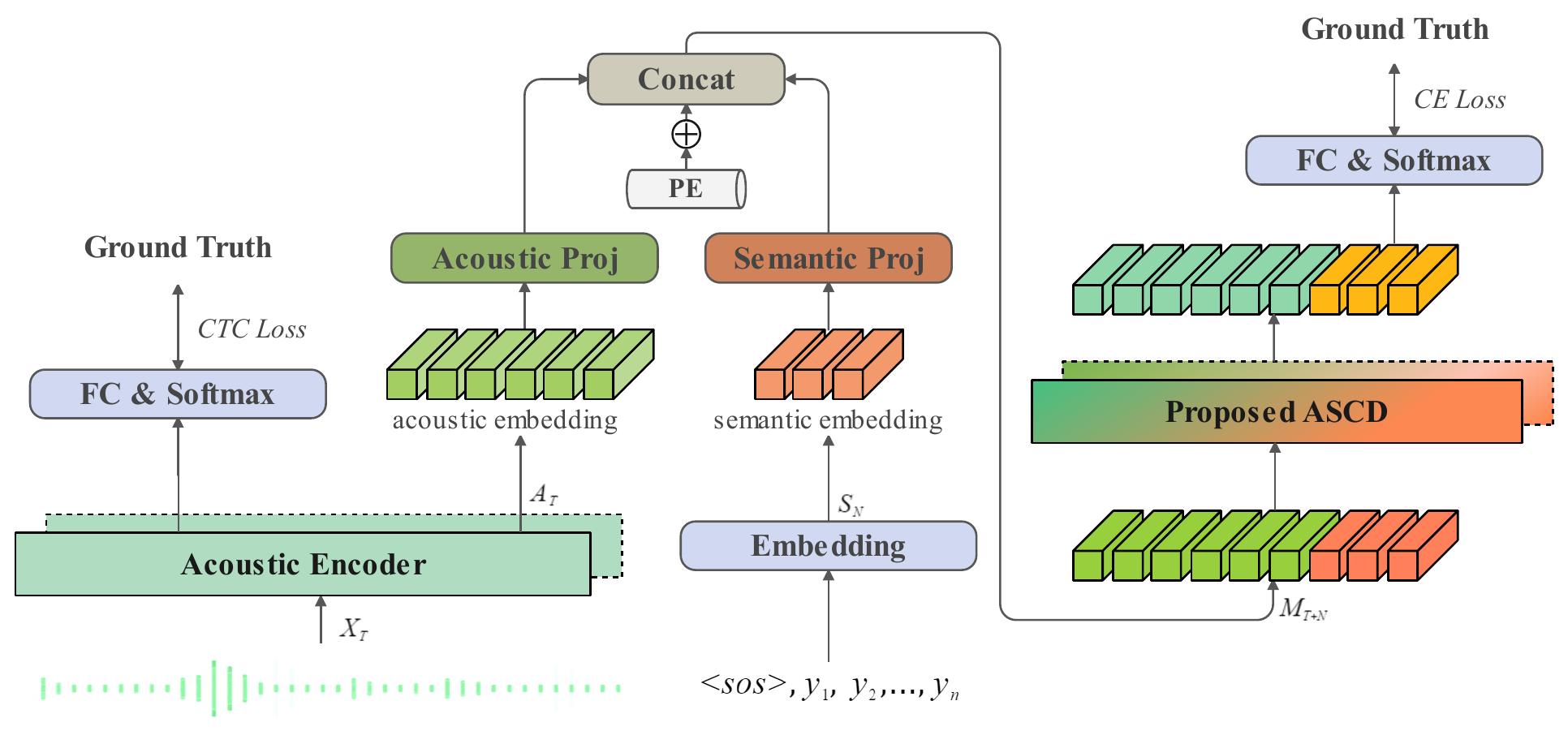}
    \vspace{-2mm}

  \caption{The pipeline of our proposed ASCD based ASR model.  The color change of the input and output of modules represents whether the feature is updated or not. Since the encoder side is not the focus of our study, the down-sampling module is omitted for brevity. (PE: positional encoding; FC: fully connected layer; $<sos>$: the start of a sentence).}
  \vspace{-4mm}
  \label{fig:arc}
\end{figure*}

In the transformer-based AED models \cite{aed1, aed2, aed3, ae4,aed5}, 
the decoder is significant for text transcription from the encoded acoustic features and the previously generated tokens.
Therefore, recently, there has been considerable attention on improving the decoder side. 
Some works \cite{nar1, nar2, nar3, nar4} focus on improving decoding speed, with the introduction of non-autoregressive decoders to generate predictions of the entire sentence.
A bidirectional decoder is proposed in \cite{bidecoder}
to capture the contextual information in forward and backward directions.
To exploit complementary information from different layers, Blockformer \cite{blockformer} is proposed with promising results. 
There are also works focus on integrating an external language model (LM) into the decoder for better domain adaptation \cite{lm1, lm2, lm3}. 
Despite numerous efforts have been made on decoder design, they overlooked the fundamental issue:
the decoder takes acoustic features and semantic features as inputs, how to leverage the complementary information of the two modalities simultaneously is crucial for obtaining high ASR performance.

The vanilla transformer decoder has two attention computational stages as shown in Fig. \ref{fig:model}(a). 
Specifically, self-attention models the semantic features in text 
to get semantic states for current time.
Differently, cross-attention identifies the most relevant encoded acoustic features
to subsequently obtain the acoustic states.
The residual connection combines the current acoustic and semantic states which are then fed into the feedforward network for further feature extraction. 
It is worth noting that the self- and cross-attention operate independently when 
producing the semantic and acoustic states. 
Considering the inherent correlation between acoustic and semantic features, their separate analysis may lead to inaccurate ASR results, as there is no acoustic information leveraged to generate semantic states.

A recent study \cite{hamilton2021parallel} 
in brain science shows that
human brain does not convert speech signals into specific tokens in a serial manner. Instead, the primary auditory cortex simply processes acoustic features where the adjacent superior temporal gyrus (STG) extracts semantic features from audio for better speech intelligibility.
Experiments verify that the primary auditory cortex and STG are activated concurrently once the cerebral cortex receives an acoustic stimuli. This indicates the parallel processing of acoustic and semantic information by the auditory cortex. 
In agreement with \cite{hamilton2021parallel}, we believe the interaction between acoustic and semantic features in the decoder stage matches their inherent correlation and complementary characteristics where their collaborative processing strategy can promote the ASR results.

Inspired by the remarkable success of multimodal fusion \cite{DBLP:journals/mms/AtreyHEK10, DBLP:journals/alife/SmithG05},
in this paper, we integrate the two fundamental modalities in ASR: the acoustic and semantic information, with a novel proposal of
\textbf{A}coustic and \textbf{S}emantic \textbf{C}ooperative \textbf{D}ecoder (ASCD).
As shown in Fig. \ref{fig:model}(b), ASCD retains only one attention module, where acoustic and semantic features are considered simultaneously and fused dynamically. 
Different with the general multimodal fusion methods, only the previous semantic information is visible due to the autoregressive decoding, thus a Causal Multimodal Mask is proposed to prevent the information leakage of the training process.
Moreover, considering the computational overhead of the models, a variant of ASCD, Semi-ASCD (S-ASCD) is proposed with less computational cost. We conduct experiments on the public available AISHELL-1 and aidatatang\_200zh datasets. Transformer \cite{DBLP:conf/icassp/DongXX18}, Conformer \cite{Gulati2020ConformerCT}, and Branchformer \cite{peng2022branchformer} are used as the encoder, respectively. In all cases, our proposed ASCD leads to improved ASR results. In particular, when using the Branchformer as the encoder, we achieve state-of-the-art results where CER equals 4.08\% and 4.33\% on the AISHELL-1 dev set and test set, respectively.
The efficacy of integrating acoustic and semantic modalities through ASCD is validated by visualizing the attention maps. 
Our contributions are listed as follows:
\begin{enumerate}
\item We propose a novel multimodal collaboration strategy for ASR, namely ASCD, which cooperatively integrates the acoustic and semantic modalities in AED decoder with improved performance.
    \item Considering the computational overhead, we propose a variant of ASCD, namely S-ASCD, that achieves comparable performance with lower computational resources.
    \item  We conduct extensive experiments on two publicly available datasets to evaluate
    our proposed ASCD model. Through visualization of the generated attention maps, we verify that collaboration of the acoustic and semantic modalities in ASCD significantly improves the ASR accuracy.
\end{enumerate}

\section{Methods}
\subsection{Overviews}
We build our model upon the  AED framework with a multi-task training strategy \cite{watanabe2017hybrid}.
The overall architecture includes a convolutional neural network based down-sampling module, an acoustic encoder and the proposed ASCD. The flow chart of the proposed model is shown in Fig. \ref{fig:arc}. Given the down-sampled speech features $\mathbf{X}_\mathrm{T}$, the acoustic encoder considers the contextual information of the speech to obtain a high-level acoustic embedding $\mathbf{A}_\mathrm{T}$, formulated as:
\begin{gather}
  \label{eq1}
  \mathbf{A}_\mathrm{T}=\mathrm{Encoder}(\mathbf{X}_\mathrm{T})
\end{gather}
where $\mathbf{A}_\mathrm{T}\in\mathbf{R}^{\mathrm{T}\times\mathrm{D_m}}$, $\mathrm{T}$ is the length of acoustic embedding and $\mathrm{D_m}$ is the model dimension.
$\mathrm{Encoder}$ is a general notation which represents the employed encoder structure. We choose  Transformer, Conformer, and Branchformer due to their widespread usage in ASR community.

We use the teacher-forcing method during training. We denote $\mathbf{Y}_\mathrm{N}=\{y_1,y_2,...,y_\mathrm{N}\}$ as the ground truth of the input speech where $\mathrm{N}$ indicates the target length. In addition, a special token $<sos>$, which indicates the beginning of a sentence, is added in $\mathbf{Y}_\mathrm{N}$ to generate the first prediction.
For the expression brevity, we still use $\mathrm{N}$ to denote the length of $\mathbf{Y}$. Then, $\mathbf{A}_\mathrm{T}$ and $\mathbf{Y}_\mathrm{N}$ are subsequently processed together and passed to ASCD for joint decoding. 

\begin{figure}[t]
  \centering
  \vspace{-0.1cm}
  \includegraphics[width=\columnwidth]{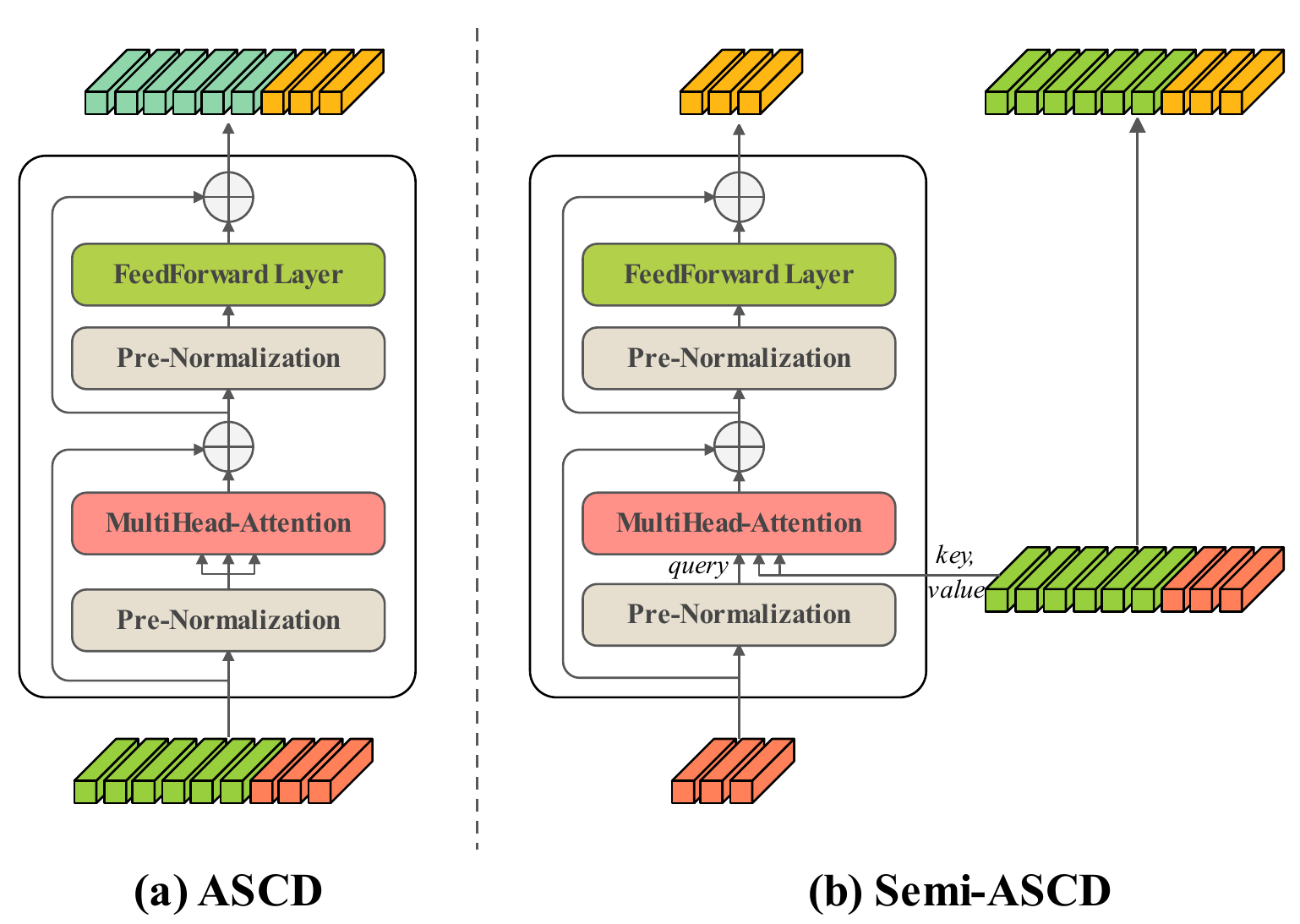}
  \vspace{-6mm}
  \caption{The structure of ASCD and Semi-ASCD layer.}
    \vspace{-2mm}
  \label{fig:ASCD}
  \vspace{-0.5cm}
\end{figure}

\vspace{-1mm}
\subsection{Acoustic and Semantic Cooperative Decoder}
\vspace{-1mm}
Given the text inputs $\mathbf{Y}_\mathrm{N}$, each token is mapped to a high-dimensional representation through the Embedding layer:
\begin{gather}
  \label{eq2}
  \mathbf{S}_\mathrm{N}=\mathrm{Embedding}(\mathbf{Y}_\mathrm{N})
\end{gather}
where $\mathbf{S}_\mathrm{N} \in \mathbf{R}^{\mathrm{T}\times\mathrm{D_e}}$ is a semantic embedding with $\mathrm{D_e}$ the output dimension.
After that, acoustic embedding $\mathbf{A}_\mathrm{T}$ and semantic embedding $\mathbf{S}_\mathrm{N}$ are projected from the space of their individual modalities to the same subspace through two fully connected acoustic projection layer and semantic projection layer, respectively. 
Finally, the two projected embeddings are concatenated along time to obtain the multimodal embedding and the positional encoding (PE) is added to provide the model with information about the relative position of each token within the input sequence. The process is formulated as:
\begin{gather}
  \label{eq3}
  \mathbf{M}_\mathrm{L}=[\mathbf{A}_\mathrm{T}\mathbf{W}^\mathrm{A};\mathbf{S}_\mathrm{N}\mathbf{W}^\mathrm{S}]\oplus\mathrm{PE}
\end{gather}
where $\mathrm{L}$ is the sum of $\mathrm{T}$ and $\mathrm{N}$. $\mathbf{W}^\mathrm{A}\in\mathbf{R}^{\mathrm{D_m}\times\mathrm{D_m}}$ and $\mathbf{W}^\mathrm{S}\in\mathbf{R}^{\mathrm{D_e}\times\mathrm{D_m}}$ are the weights of the acoustic and semantic projection layers, respectively. $\oplus$ denotes addition operations.

The obtained multimodal embedding $\mathbf{M}_\mathrm{L}$ is fed into the ASCD with each layer structured as shown in Fig. \ref{fig:ASCD}(a), consisting of only a multi-head attention module and FFN module, both with pre-normalization and shortcut connection.
In ASCD, each element of the semantic features interacts with the both acoustic features and the previous predicted semantic features through multi-head attention, such that the relevant features of both modalities are simultaneously involved in reasoning about the current semantic state. At the same time, the acoustic features also interacts internally, and in the same attention computation process as the interaction between the two modalities, so that the acoustic features are updated guided by the semantic information. The process of ASCD can be written as follows:

\begin{equation}
    \label{eq4}
    \begin{split}
    &\hat{\mathbf{M}}_\mathrm{L}=\mathrm{LayerNorm}(\mathbf{M}_\mathrm{L}) \\
    &\mathbf{M}'_\mathrm{L}=\mathrm{MHA}(\mathbf{q},\mathbf{k},\mathbf{v}=\hat{\mathbf{M}}_\mathrm{L})+\mathbf{M}_\mathrm{L} \\
    &\hat{\mathbf{M}}'_\mathrm{L}=\mathrm{LayerNorm}(\mathbf{M}'_\mathrm{L}) \\
    &\mathbf{M}''_\mathrm{L}=\mathrm{FFN}(\hat{\mathbf{M}}'_\mathrm{L})+\mathbf{M}'_\mathrm{L} \\
    \end{split}
\end{equation}
where $\mathrm{MHA}$ represents MultiHead Attention and $\mathrm{FFN}$ represents FeedForward Layer.

However, compared to the semantic-only input of vanilla decoder, the multimodal embedding $\mathbf{M}_\mathrm{L}$ has an additional length $\mathrm{T}$ of acoustic embedding, which is involved in the computation of all components in each ASCD layer and therefore brings additional computation overhead. Therefore, we propose an alternative cooperative method, S-ASCD. As shown in Fig. \ref{fig:ASCD}(b), only the semantic part $\mathbf{S}_\mathrm{N}$ is computed in all components in S-ASCD , while the multimodal embedding $\mathbf{M}_\mathrm{L}$ is only participated in the multi-head attention as key and value. In this way, the reasoning of semantic states still considers the information of both modalities, but the acoustic features is no longer updated. With the high computational overhead removed, the time complexity of the algorithm is decreased from $O((\mathrm{T}+\mathrm{N})^2)$ to $O(\mathrm{N}(\mathrm{T}+\mathrm{N}))$, which is consistent with vanilla decoder. The formula process follows:

\begin{equation}
    \label{eq5}
        \vspace{-0.1cm}
    \begin{split}
    &\hat{\mathbf{S}}_\mathrm{N}=\mathrm{LayerNorm}(\mathbf{S}_\mathrm{N}) \\
    &\mathbf{S}'_\mathrm{N}=\mathrm{MHA}(\mathbf{q}=\hat{\mathbf{S}}_\mathrm{N};\mathbf{k},\mathbf{v}=\mathbf{M}_\mathrm{L})+\mathbf{S}_\mathrm{N} \\
    &\hat{\mathbf{S}}'_\mathrm{N}=\mathrm{LayerNorm}(\mathbf{S}'_\mathrm{N}) \\
    &\mathbf{S}''_\mathrm{N}=\mathrm{FFN}(\hat{\mathbf{S}}'_\mathrm{N})+\mathbf{S}'_\mathrm{N} \\
    & \mathbf{M}'_\mathrm{L}=[\mathbf{A}_\mathrm{T};\mathbf{S}''_\mathrm{N}]
    \end{split}
\end{equation}
where $\mathbf{M}'_\mathrm{L}$ is the updated multimodal embedding used for the next layer. For both ASCD and S-ASCD, we only constrain the semantic features of the output with a cross-entropy loss.

\begin{figure}[t]
  \centering
  \includegraphics[width=\linewidth]{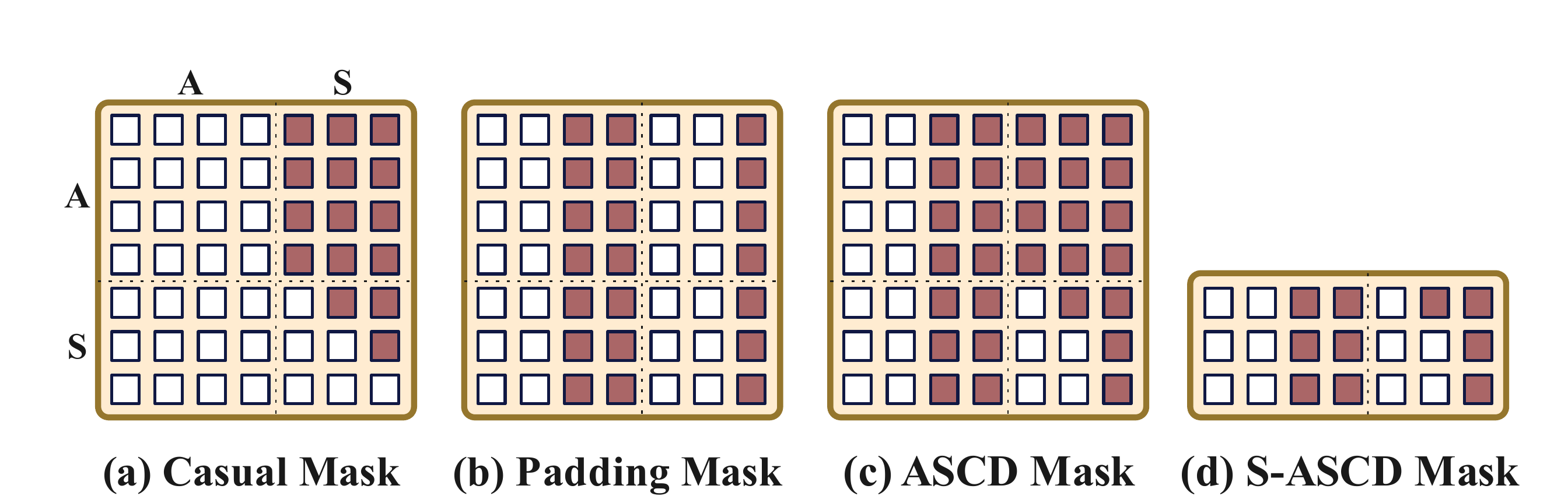}
  \caption{Details of the mask strategy for ASCD and S-ASCD. The red block represents the masked position which will be assigned the $-inf$ value.}
  \label{fig:mask}
    \vspace{-10mm}
\end{figure}

\vspace{-1mm}
\subsection{Masking Strategies}
\vspace{-1mm}
In the attention computation of ASCD, there are four types of interaction between modalities: acoustic-to-acoustic (A-A), acoustic-to-semantic (A-S), semantic-to-acoustic (S-A), and semantic-to-semantic (S-S). To ensure that each element in the S-S interaction is only influenced by the decoded tokens that precede it, an upper triangular mask is employed to shield information from future time steps, effectively avoiding the possibility of any future information leakage. However, the precautionary measures do not end there. In order to prevent the acoustic features from carrying semantic information of future moments, which can cause an unintended breach of information in the subsequent layer, a mask matrix is also necessary for the interaction of S-A. As illustrated in Fig. \ref{fig:mask}(a), the integration of these two masks produces a Causal Mask that effectively forestalls any potential leakage of semantic information from future time steps.

To maintain the length consistency in each batch, we apply zero-padding to each speech signal and target label during the pre-processing phase.
To prevent the model from using the padded region during training, a Padding Mask, as shown in Fig. \ref{fig:mask}(b), is applied to conceal it which not only ensures consistency but also accelerates the model's convergence. The union of the Causal Mask and Padding Mask results in the ASCD mask as delineated in Fig. \ref{fig:mask}(c). Moreover, the S-ASCD mask is composed of the A-A and A-S components of the ASCD mask as depicted in the Fig. \ref{fig:mask}(d). 

\begin{table}[t] 
    \caption{Results on the AISHELL-1 dataset. In our Implementations, CTC loss is used on the Conformer and Branchformer, but not on the Transformer.}
    \vspace{-2mm}
    \centering
    \small
    \label{table1}
    \setlength{\tabcolsep}{3.8mm}{
    \begin{tabular}{l cc c}
    \toprule
    \multirow{2}{*}{\textbf{Models}} & \multicolumn{2}{c}{\textbf{CER (\%)}}  & \multirow{2}{*}{\textbf{Params (M)}} \\
    & \makecell{Dev} & \makecell{Test} \\
    \midrule
        \midrule
    \multicolumn{4}{l}{\textit{Published on the ESPnet} \cite{DBLP:conf/interspeech/WatanabeHKHNUSH18}} \\
    \midrule
     Transformer & 5.9 & 6.4 &30.4 \\
     Conformer & 4.3 & 4.6 & 46.2 \\
     Branchformer & \textbf{4.2} & \textbf{4.4} & 45.4 \\
    \midrule
        \midrule

    \multicolumn{4}{l}{\textit{Ours (Implemented on the ESPnet)}} \\
        \midrule
    
    Transformer & 6.07 & 6.53 & 29.26 \\
    \quad+ ASCD & 5.37 & 5.82 & \multirow{2}{*}{27.81}\\
    \quad+ S-ASCD & \textbf{5.36} & \textbf{5.80} \\
    \midrule
    Conformer & 4.28 & 4.69 & 46.25 \\
    \quad+ ASCD & 4.23 & \textbf{4.52} & \multirow{2}{*}{44.80}  \\
    \quad+ S-ASCD & \textbf{4.22} & 4.54  \\
    \midrule
    Branchformer & 4.12  & 4.47  & 45.43 \\
    \quad+ ASCD & 4.08  & \textbf{4.33}  & \multirow{2}{*}{43.98} \\
    \quad+ S-ASCD & \textbf{4.07} & 4.38  \\ 
    \bottomrule
    \end{tabular}}
    \vspace{-0.4cm}
\end{table}

\vspace{-2mm}
\section{Experiments}
\subsection{Experimental Setup}
\vspace{-1mm}
We conduct experiments on  AISHELL-1 \cite{Aishell} and aidatatang\_200zh$\footnote{https://www.datatang.com}$. To represent the input vectors, we employ a sequence of 80-dimensional log-Mel filter bank with 3-dimensional pitch features. Each frame is computed with a window length of 25 ms and a shift of 10 ms and normalized using Global CMVN \cite{DBLP:journals/speech/ViikkiL98}. To enhance the robustness of the model, we adopt the speed perturbation \cite{sp} and SpecAugment \cite{spec} during the training process. Our experiments are conducted using the NVIDIA Tesla V100 GPUs with the same configuration as the official ESPnet$\footnote{https://github.com/espnet/espnet/tree/master/egs2}$. We employ the widely used  character error rate (CER) metric for evaluation.

\begin{table}[h]
\vspace{-0.1cm}
    \caption{Results on the aidatang\_200zh dataset.}
        \vspace{-2mm}
    \centering
    \small
    \label{table2}
      \setlength{\abovecaptionskip}{-10.mm}

    \renewcommand\arraystretch{1.0}
    \setlength{\tabcolsep}{4mm}{
    \begin{tabular}{lccc}
    \toprule
   \multirow{2}{*}{\textbf{Models}} & \multicolumn{2}{c}{\textbf{CER (\%)}} & \multirow{2}{*}{\textbf{Params (M)}}\\
    & \makecell{Dev} & \makecell{Test} \\
    \midrule
    Conformer & 3.65 & 4.31& 45.98 \\ 
     \quad + ASCD & 3.58& \textbf{4.18} & \multirow{2}{*}{44.53}\\
     \quad + S-ASCD & \textbf{3.54} & 4.19 \\
    \bottomrule
    \end{tabular}}
    \vspace{-0.4cm}
\end{table}

\vspace{-1mm}
\subsection{Results}
\vspace{-1mm}
Table \ref{table1} shows the results on the AISHELL-1 dataset. We first report the public results on the ESPnet, where the Conformer and Branchformer results represent the state-of-the-art public level. Subsequently, we present the baseline results obtained through our independent implementation of the three models, followed by the results obtained with the incorporation of ASCD. It is noteworthy that in the case of the Transformer model, we intentionally abstain from incorporating an additional CTC loss \cite{watanabe2017hybrid} for encoder supervision. This deliberate omission is aimed at generating an acoustic embedding of insufficient quality, thereby facilitating a more accurate evaluation of the effectiveness of ASCD in enhancing the performance of the model. As a result, ASCD achieves 5.37\% and 5.82\% CER results on the dev and test sets, respectively, and S-ASCD achieves 5.36\% and 5.80\% CER results, with a relative CER reduction (CERR) of 11.18\% on the test set compared to the transformer baseline. For Conformer, ASCD and S-ASCD achieve 3.62\% and 3.20\% CERR on the test set, respectively. When using Branchformer as encoder, ASCD achieves a state-of-the art results of 4.08\%$\backslash$4.33\% CER on the dev$\backslash$test set, and S-ASCD achieves 4.07\%$\backslash$4.38\% CER on the dev$\backslash$test set. It is observed that the results of ASCD is marginally superior to S-ASCD. This advantage may be attributed to the synchronized update of acoustic features. The outstanding results demonstrate the importance of simultaneously considering acoustic and semantic features for generating accurate semantic states. Additionally, as each layer of ASCD comprises one less attention layer than the conventional decoder layer, the former inherently possesses a lower number of parameters.

The results on the aidatatang\_200zh dataset are shown in Table \ref{table2}. ASCD and S-ASCD achieve 3.58\%$\backslash$4.18\% and 3.54\%$\backslash$4.19\% CER on the dev$\backslash$test sets, respectively. Compared to the previous best result on the test set, a relative CERR of 3.02\% is achieved.

\begin{figure}[t]
 \vspace{-0.3cm}
  \includegraphics[width=\linewidth]{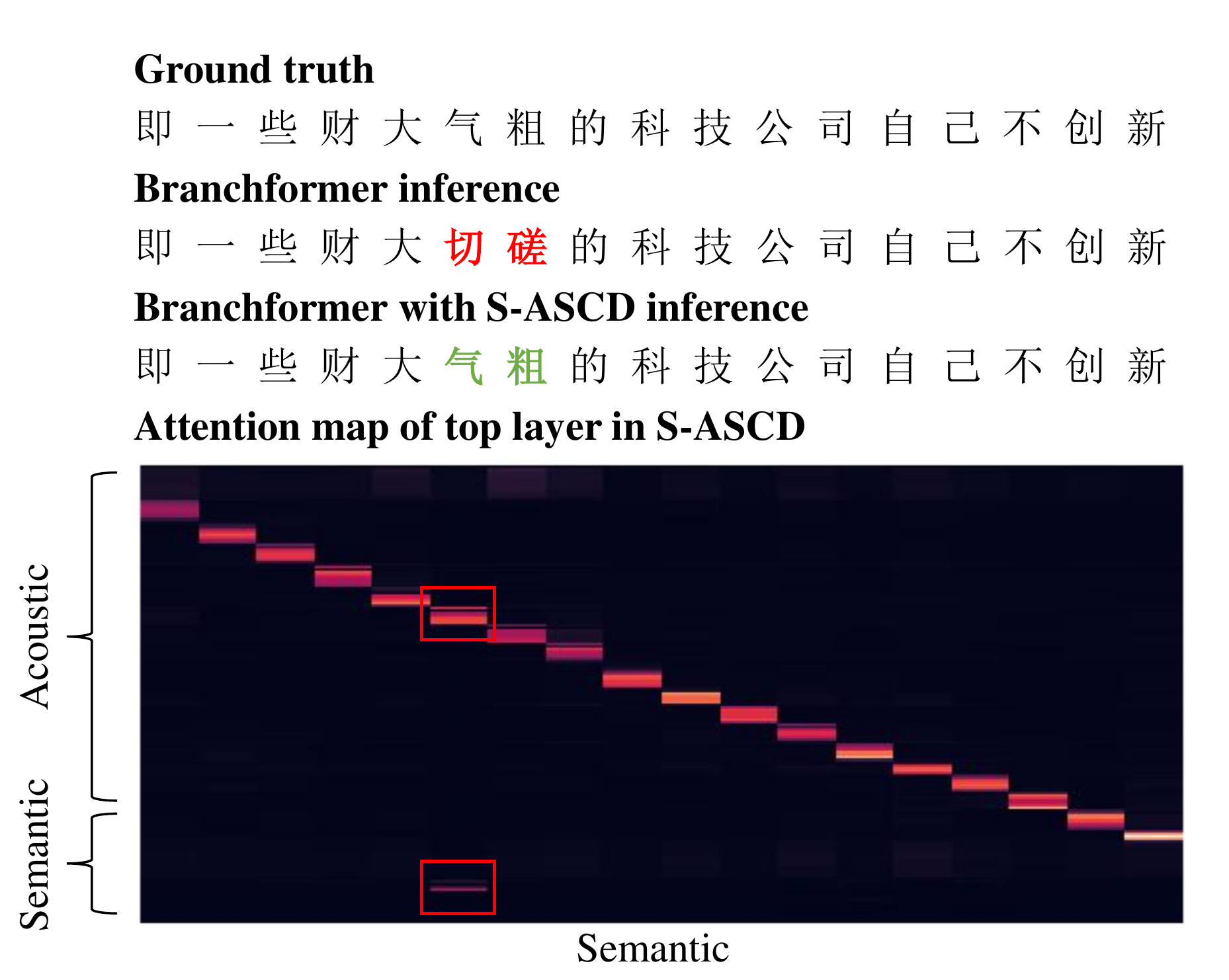}
  \caption{A decoding example 
  of the AISHELL-1 test set. Red denotes incorrectly predicted tokens, while green represents tokens correctly predicted by considering both acoustic and semantic features (red box in the attention map).}
  \label{fig:vis}
    \vspace{-0.6cm}
\end{figure}

\vspace{-1mm}
\subsection{Analysis}
\vspace{-1mm}
In order to have a comprehensive understanding of our proposed ASCD, we conduct detailed visualizations of specific samples. As shown in Fig. \ref{fig:vis}, the topmost row represents the ground truth sequence of characters, whereas the second and third rows correspond to the output hypotheses generated by the baseline model and our S-ASCD proposal, respectively. The bottom row is the attention map generated by the top layer of S-ASCD. In this instance, Branchformer erroneously recognizes a result with a similar pronunciation. However, S-ASCD produces a correct prediction by simultaneously attending to both acoustic and semantic features as shown in the attention map. This becomes evident that cooperatively decoding with acoustic and semantic features leads to more accurate results.

\vspace{-2mm}
\section{Conclusion}
This study takes a fresh perspective on ASR task by adopting a multimodal fusion approach to effectively leverage both acoustic and semantic features for generating semantic states. Considering the computational cost, we propose a variant of ASCD, namely S-ASCD, which consumes less computation. The proposed methods achieve state-of-the-art results on two public datasets. Furthermore, the effectiveness of considering both acoustic and semantic features simultaneously is demonstrated through visualization analysis. In future work, we intend to explore the combination of ASCD and non-autoregressive methods, as well as investigate more efficient interaction patterns between the acoustic and semantic features.

\textbf{Acknowledgements:} The research is supported by National Key Research and Development Program of China (2020AAA0109700), National Natural Science Foundation of China (62076024, 62006018, U22B2055).

\bibliographystyle{IEEEtran}
\bibliography{mybib}

\begin{thebibliography}{10}
\providecommand{\url}[1]{#1}
\csname url@samestyle\endcsname
\providecommand{\newblock}{\relax}
\providecommand{\bibinfo}[2]{#2}
\providecommand{\BIBentrySTDinterwordspacing}{\spaceskip=0pt\relax}
\providecommand{\BIBentryALTinterwordstretchfactor}{4}
\providecommand{\BIBentryALTinterwordspacing}{\spaceskip=\fontdimen2\font plus
\BIBentryALTinterwordstretchfactor\fontdimen3\font minus
  \fontdimen4\font\relax}
\providecommand{\BIBforeignlanguage}[2]{{%
\expandafter\ifx\csname l@#1\endcsname\relax
\typeout{** WARNING: IEEEtran.bst: No hyphenation pattern has been}%
\typeout{** loaded for the language `#1'. Using the pattern for}%
\typeout{** the default language instead.}%
\else
\language=\csname l@#1\endcsname
\fi
#2}}
\providecommand{\BIBdecl}{\relax}
\BIBdecl

\bibitem{ctc}
A.~Graves, S.~Fern{\'{a}}ndez, F.~J. Gomez, and J.~Schmidhuber, ``Connectionist
  temporal classification: labelling unsegmented sequence data with recurrent
  neural networks,'' in \emph{Machine Learning, Proceedings of the Twenty-Third
  International Conference {ICML}}, vol. 148, 2006, pp. 369--376.

\bibitem{graves2012sequence}
A.~Graves, ``Sequence transduction with recurrent neural networks,''
  \emph{arXiv preprint arXiv:1211.3711}, 2012.

\bibitem{asr1}
T.~N. Sainath, Y.~He, B.~Li, A.~Narayanan, R.~Pang \emph{et~al.}, ``A streaming
  on-device end-to-end model surpassing server-side conventional model quality
  and latency,'' in \emph{International Conference on Acoustics, Speech and
  Signal Processing, {ICASSP}}, 2020, pp. 6059--6063.

\bibitem{asr2}
J.~Li, R.~Zhao, Z.~Meng, Y.~Liu, W.~Wei \emph{et~al.}, ``Developing {RNN-T}
  models surpassing high-performance hybrid models with customization
  capability,'' in \emph{21st Annual Conference of the International Speech
  Communication Association, {INTERSPEECH}}, 2020, pp. 3590--3594.

\bibitem{aed1}
C.~Chiu, T.~N. Sainath, Y.~Wu, R.~Prabhavalkar, P.~Nguyen \emph{et~al.},
  ``State-of-the-art speech recognition with sequence-to-sequence models,'' in
  \emph{International Conference on Acoustics, Speech and Signal Processing,
  {ICASSP}}, 2018, pp. 4774--4778.

\bibitem{vaswani2017attention}
A.~Vaswani, N.~Shazeer, N.~Parmar, J.~Uszkoreit, L.~Jones, A.~N. Gomez,
  {\L}.~Kaiser, and I.~Polosukhin, ``Attention is all you need,''
  \emph{Advances in neural information processing systems}, vol.~30, 2017.

\bibitem{aed2}
J.~Chorowski, D.~Bahdanau, D.~Serdyuk, K.~Cho, and Y.~Bengio, ``Attention-based
  models for speech recognition,'' in \emph{Annual Conference on Neural
  Information Processing Systems, {NIPS}}, 2015, pp. 577--585.

\bibitem{aed3}
D.~Bahdanau, J.~Chorowski, D.~Serdyuk, P.~Brakel, and Y.~Bengio, ``End-to-end
  attention-based large vocabulary speech recognition,'' in \emph{International
  Conference on Acoustics, Speech and Signal Processing, {ICASSP}}, 2016, pp.
  4945--4949.

\bibitem{ae4}
W.~Chan, N.~Jaitly, Q.~V. Le, and O.~Vinyals, ``Listen, attend and spell: {A}
  neural network for large vocabulary conversational speech recognition,'' in
  \emph{International Conference on Acoustics, Speech and Signal Processing,
  {ICASSP}}, 2016, pp. 4960--4964.

\bibitem{aed5}
S.~Karita, X.~Wang, S.~Watanabe, T.~Yoshimura, W.~Zhang \emph{et~al.}, ``A
  comparative study on transformer vs {RNN} in speech applications,'' in
  \emph{Automatic Speech Recognition and Understanding Workshop, {ASRU}}, 2019,
  pp. 449--456.

\bibitem{nar1}
Y.~Bai, J.~Yi, J.~Tao, Z.~Tian, Z.~Wen \emph{et~al.}, ``Listen attentively, and
  spell once: Whole sentence generation via a non-autoregressive architecture
  for low-latency speech recognition,'' in \emph{21st Annual Conference of the
  International Speech Communication Association, Virtual Event,
  {INTERSPEECH}}, 2020, pp. 3381--3385.

\bibitem{nar2}
Y.~Higuchi, S.~Watanabe, N.~Chen, T.~Ogawa, and T.~Kobayashi, ``Mask {CTC:}
  non-autoregressive end-to-end {ASR} with {CTC} and mask predict,'' in
  \emph{21st Annual Conference of the International Speech Communication
  Association, {INTERSPEECH}}, 2020, pp. 3655--3659.

\bibitem{nar3}
C.~Zhang, Y.~Liu, T.~Zhang, S.~Chen, F.~Chen, and X.~Yin, ``Non-autoregressive
  transformer with unified bidirectional decoder for automatic speech
  recognition,'' in \emph{International Conference on Acoustics, Speech and
  Signal Processing, {ICASSP}}, 2022, pp. 6527--6531.

\bibitem{nar4}
Z.~Gao, S.~Zhang, I.~McLoughlin, and Z.~Yan, ``Paraformer: Fast and accurate
  parallel transformer for non-autoregressive end-to-end speech recognition,''
  in \emph{23rd Annual Conference of the International Speech Communication
  Association, {INTERSPEECH}}, 2022, pp. 2063--2067.

\bibitem{bidecoder}
X.~Chen, S.~Zhang, D.~Song, P.~Ouyang, and S.~Yin, ``Transformer with
  bidirectional decoder for speech recognition,'' in \emph{Interspeech 2020,
  21st Annual Conference of the International Speech Communication Association,
  {INTERSPEECH}}, 2020, pp. 1773--1777.

\bibitem{blockformer}
X.~Ren, H.~Zhu, L.~Wei, M.~Wu, and J.~Hao, ``Improving mandarin speech
  recogntion with block-augmented transformer,'' \emph{arXiv preprint
  arXiv:2207.11697}, 2022.

\bibitem{lm1}
Z.~Meng, S.~Parthasarathy, E.~Sun, Y.~Gaur, N.~Kanda \emph{et~al.}, ``Internal
  language model estimation for domain-adaptive end-to-end speech
  recognition,'' in \emph{Spoken Language Technology Workshop, {SLT}}, 2021,
  pp. 243--250.

\bibitem{lm2}
E.~Tsunoo, Y.~Kashiwagi, C.~P. Narisetty, and S.~Watanabe, ``Residual language
  model for end-to-end speech recognition,'' in \emph{23rd Annual Conference of
  the International Speech Communication Association, {INTERSPEECH}}, 2022, pp.
  3899--3903.

\bibitem{lm3}
Z.~Meng, N.~Kanda, Y.~Gaur, S.~Parthasarathy, E.~Sun \emph{et~al.}, ``Internal
  language model training for domain-adaptive end-to-end speech recognition,''
  in \emph{International Conference on Acoustics, Speech and Signal Processing,
  {ICASSP}}, 2021, pp. 7338--7342.

\bibitem{hamilton2021parallel}
L.~S. Hamilton, Y.~Oganian, J.~Hall, and E.~F. Chang, ``Parallel and
  distributed encoding of speech across human auditory cortex,'' \emph{Cell},
  vol. 184, no.~18, pp. 4626--4639, 2021.

\bibitem{DBLP:journals/mms/AtreyHEK10}
P.~K. Atrey, M.~A. Hossain, A.~El{-}Saddik, and M.~S. Kankanhalli, ``Multimodal
  fusion for multimedia analysis: a survey,'' \emph{Multim. Syst.}, vol.~16,
  no.~6, pp. 345--379, 2010.

\bibitem{DBLP:journals/alife/SmithG05}
L.~B. Smith and M.~Gasser, ``The development of embodied cognition: Six lessons
  from babies,'' \emph{Artif. Life}, vol.~11, no. 1-2, pp. 13--29, 2005.

\bibitem{DBLP:conf/icassp/DongXX18}
L.~Dong, S.~Xu, and B.~Xu, ``Speech-transformer: {A} no-recurrence
  sequence-to-sequence model for speech recognition,'' in \emph{International
  Conference on Acoustics, Speech and Signal Processing, {ICASSP}}, 2018, pp.
  5884--5888.

\bibitem{Gulati2020ConformerCT}
A.~Gulati, J.~Qin, C.-C. Chiu, N.~Parmar, Y.~Zhang, J.~Yu, W.~Han, S.~Wang,
  Z.~Zhang, Y.~Wu, and R.~Pang, ``Conformer: Convolution-augmented transformer
  for speech recognition,'' \emph{ISCA}, vol. abs/2005.08100, 2020.

\bibitem{peng2022branchformer}
Y.~Peng, S.~Dalmia, I.~Lane, and S.~Watanabe, ``Branchformer: Parallel
  mlp-attention architectures to capture local and global context for speech
  recognition and understanding,'' in \emph{ICML}.\hskip 1em plus 0.5em minus
  0.4em\relax PMLR, 2022, pp. 17\,627--17\,643.

\bibitem{watanabe2017hybrid}
S.~Watanabe, T.~Hori, S.~Kim, J.~R. Hershey, and T.~Hayashi, ``Hybrid
  ctc/attention architecture for end-to-end speech recognition,'' \emph{IEEE
  Journal of Selected Topics in Signal Processing}, vol.~11, no.~8, pp.
  1240--1253, 2017.

\bibitem{DBLP:conf/interspeech/WatanabeHKHNUSH18}
S.~Watanabe, T.~Hori, S.~Karita, T.~Hayashi, J.~Nishitoba \emph{et~al.},
  ``Espnet: End-to-end speech processing toolkit,'' in \emph{19th Annual
  Conference of the International Speech Communication Association,
  {Interspeech}}, B.~Yegnanarayana, Ed., pp. 2207--2211.

\bibitem{Aishell}
H.~Bu, J.~Du, X.~Na, B.~Wu, and H.~Zheng, ``{AISHELL-1:} an open-source
  mandarin speech corpus and a speech recognition baseline,'' in \emph{20th
  Conference of the Oriental Chapter of the International Coordinating
  Committee on Speech Databases and Speech {I/O} Systems and Assessment,
  {O-COCOSDA}}, 2017, pp. 1--5.

\bibitem{DBLP:journals/speech/ViikkiL98}
O.~Viikki and K.~Laurila, ``Cepstral domain segmental feature vector
  normalization for noise robust speech recognition,'' \emph{Speech Commun.},
  vol.~25, no. 1-3, pp. 133--147, 1998.

\bibitem{sp}
T.~Ko, V.~Peddinti, D.~Povey, and S.~Khudanpur, ``Audio augmentation for speech
  recognition,'' in \emph{16th Annual Conference of the International Speech
  Communication Association, {INTERSPEECH}}, 2015, pp. 3586--3589.

\bibitem{spec}
D.~S. Park, W.~Chan, Y.~Zhang, C.~Chiu, B.~Zoph \emph{et~al.}, ``Specaugment:
  {A} simple data augmentation method for automatic speech recognition,'' in
  \emph{20th Annual Conference of the International Speech Communication
  Association, {INTERSPEECH}}, 2019, pp. 2613--2617.

\end{thebibliography}

\end{document}